\documentclass[runningheads]{llncs}
\usepackage{eccv}
\usepackage{eccvabbrv}
\usepackage{graphicx}
\usepackage{booktabs}
\usepackage{tabularx}
\usepackage[accsupp]{axessibility}  
\usepackage{xcolor}
\usepackage{colortbl}
\usepackage{siunitx}
\usepackage{algorithmicx,algorithm}
\usepackage[noend]{algpseudocode}

\definecolor{topcolor}{rgb}{1, 0.8, 0.8} 
\definecolor{secondcolor}{rgb}{1, 0.87, 0.7} 
\definecolor{thirdcolor}{rgb}{1, 1, 0.8}

\usepackage{hyperref}

\usepackage{orcidlink}
\usepackage{nicematrix,makecell}

\begin{document}

\title{Deceptive-NeRF/3DGS: Diffusion-Generated Pseudo-Observations for High-Quality Sparse-View Reconstruction }
\titlerunning{Deceptive-NeRF/3DGS}

\author{Xinhang Liu\inst{1}\orcidlink{0009-0003-0494-4877} \and
Jiaben Chen\inst{2}\orcidlink{0000-0002-2252-1419} \and
Shiu-Hong Kao\inst{1}\orcidlink{0000-0002-8367-8487} \and
Yu-Wing Tai\inst{3}\orcidlink{0000-0002-3148-0380}   \and
Chi-Keung Tang\inst{1}\orcidlink{0000-0001-7155-2919}}

\authorrunning{X.~Liu et al.}

\institute{The Hong Kong University of Science and Technology \and
University of California, San Diego \and
Dartmouth College\\
\email{\{xliufe,skao\}@connect.ust.hk, jic088@ucsd.edu, yuwing@gmail.com, cktang@cs.ust.hk}}

\maketitle
\setcounter{footnote}{0}
\begin{abstract}
Novel view synthesis via Neural Radiance Fields (NeRFs) or 3D Gaussian Splatting (3DGS) typically necessitates dense observations with hundreds of input images to circumvent artifacts. 
We introduce {\sc Deceptive-NeRF/3DGS}\footnote{In harmonic progression, a Deceptive Cadence may disrupt expectations of chord progression but enriches the emotional expression of the music. Our Deceptive-X, where ``X'' can be NeRF, 3DGS, or a pertinent 3D reconstruction framework—counters overfitting to sparse input views by densely synthesizing consistent pseudo-observations, enriching the original sparse inputs by fivefold to tenfold.} to enhance sparse-view reconstruction with only a limited set of input images, by leveraging a diffusion model pre-trained from multiview datasets. 
Different from using diffusion priors to regularize representation optimization, our method directly uses diffusion-generated images to train NeRF/3DGS as if they were real input views.
Specifically, we propose a \emph{deceptive diffusion model} turning noisy images rendered from few-view reconstructions into high-quality photorealistic \emph{pseudo-observations}.
To resolve consistency among pseudo-observations and real input views, we develop an uncertainty measure to guide the diffusion model's generation. 
Our system progressively incorporates diffusion-generated pseudo-observations into the training image sets, ultimately densifying the sparse input observations by 5 to 10 times.
Extensive experiments across diverse and challenging datasets validate that our approach outperforms existing state-of-the-art methods and is capable of synthesizing novel views with super-resolution in the few-view setting.
Project page: \href{https://xinhangliu.com/deceptive-nerf-3dgs}{https://xinhangliu.com/deceptive-nerf-3dgs}.
\end{abstract}

\begin{figure}[t]
\begin{center}
    \includegraphics[width=\linewidth]{ 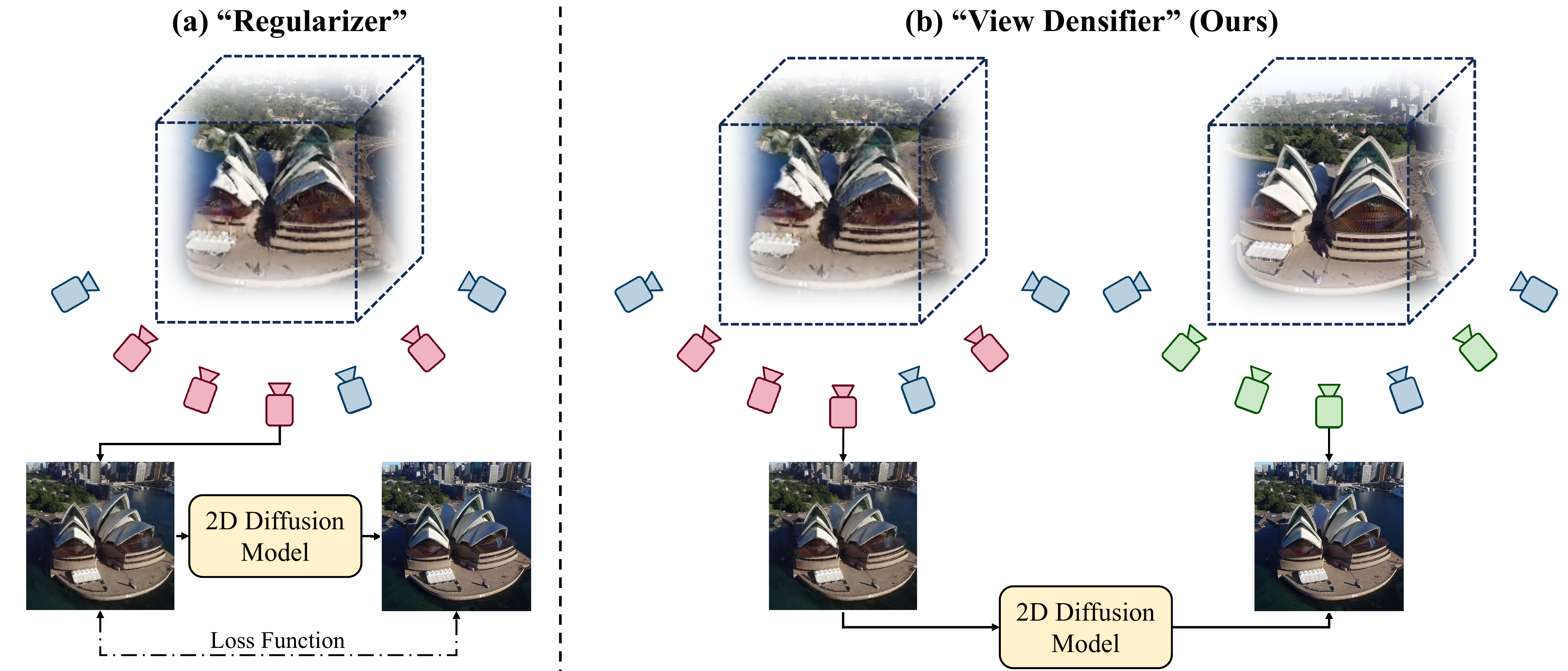}
\end{center}
\caption{\textbf{Different approaches to applying 2D diffusion priors in few-view 3D reconstruction.}  \textbf{(a)} With only a few \includegraphics[width=0.03\linewidth]{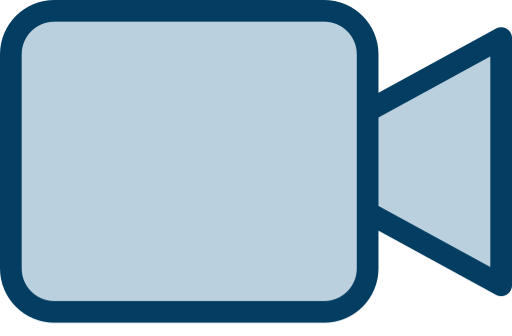} \color{blue} input images \color{black},  an intuition is to utilize the 2D diffusion model as a ``scorer'' for synthesized \includegraphics[width=0.03\linewidth] {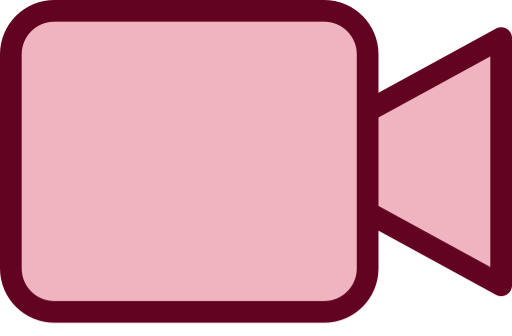} \color[rgb]{0.75,0, 0} novel views \color{black}, regularizing NeRF/3DGS training.
\textbf{(b)} Instead, our approach densifies input views by generating dense \includegraphics[width=0.03\linewidth]{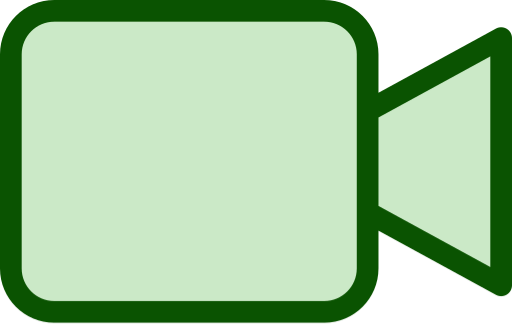} \color[rgb]{0,0.5, 0} pseudo-observations \color{black} that are consistent with the given inputs to progressively enhance reconstruction as shown. This approach avoids the need to infer the diffusion model at every training step, thereby offering the advantage of nearly tenfold faster training speed.}
\label{fig:overall_method}
\end{figure}

\begin{figure}[t]
\begin{center}
    \includegraphics[width=0.95\linewidth]{ 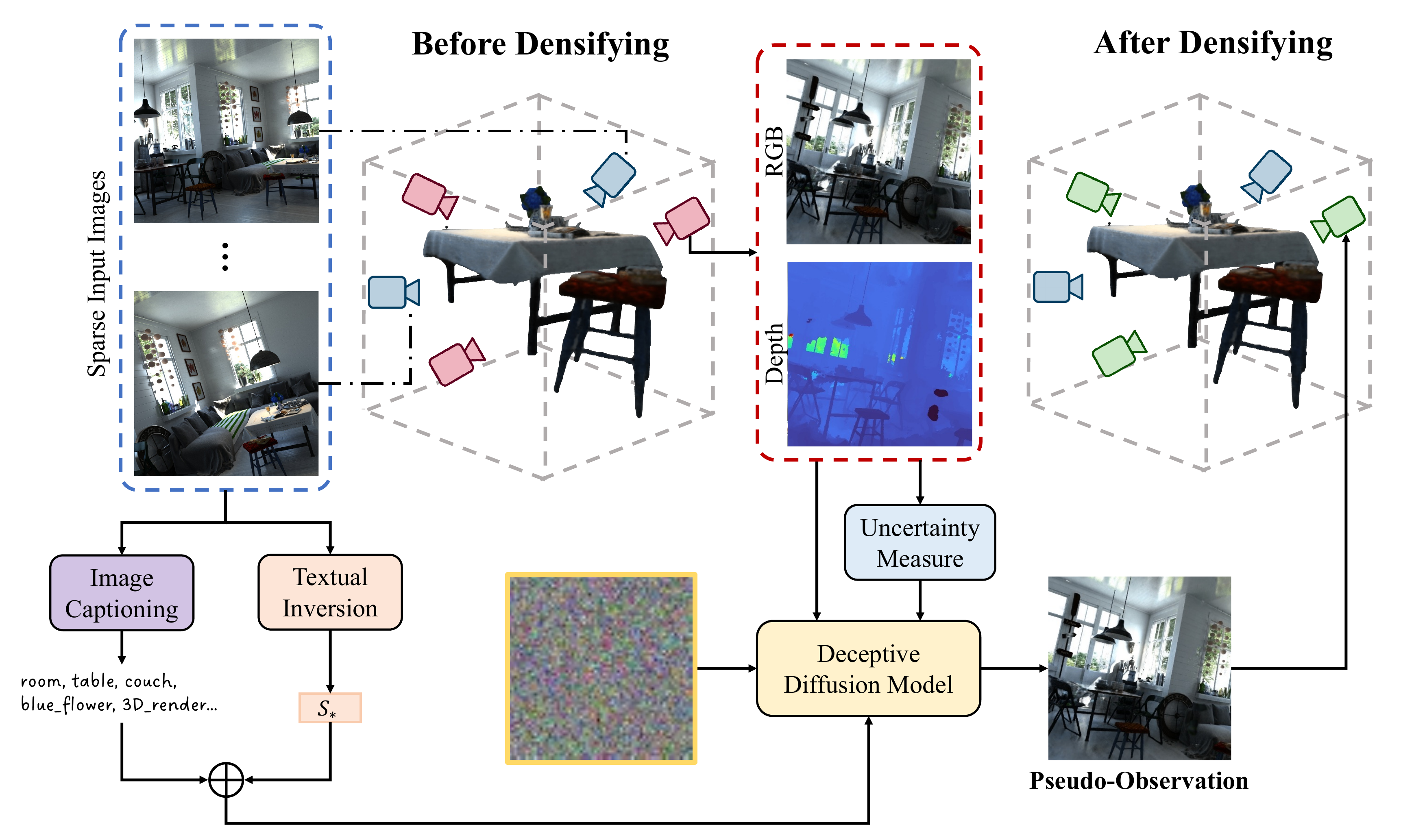}
\end{center}
\caption{\textbf{Overview of Deceptive-NeRF/3DGS.} 1) Given a sparse set of \includegraphics[width=0.03\linewidth]{figures/cam_blue.png} \color{blue}input images \color{black} with camera poses, we train NeRF/3DGS to render \color[rgb]{0.75,0, 0}\includegraphics[width=0.03\linewidth]{figures/cam_red.png} coarse novel view \color{black} images and depth maps. 
2) Our deceptive diffusion model enhances RGB-D images from coarse reconstruction, along with a novel uncertainty measure, to generate \color[rgb]{0,0.5, 0}\includegraphics[width=0.03\linewidth]{figures/cam_green.png} pseudo-observations \color{black} from corresponding viewpoints. 
3) We continue training NeRF/3DGS using both input images (real) and pseudo-observations (fake) and repeat the aforementioned process to get our final reconstruction.
}
\label{fig:method}
\end{figure}

\section{Introduction}
Recent novel view synthesis approaches like Neural Radiance Fields (NeRFs)~\cite{nerf} and 3D Gaussian Splatting~\cite{3dgs} have achieved unprecedented results. 
However, without exception, all these pipelines~\cite{chen2022tensorf,muller2022instant,barron2023zip} require a large number of training views to produce visually pleasing results and are prone to generating severe artifacts when dealing with sparse observations. 
This issue can hamper their further consumer-level usage, where casual data collection by lay users is often made up of only sparse image capture using mobile devices.

To adapt NeRF to the few-view setting, some methods leverage large-scale datasets comprising various scenes to pre-train their models and inject prior knowledge~\cite{yu2020pixelnerf, mvsnerf, SRF, jang2021codenerf,johari2022geonerf}.
A larger portion of methods~\cite{Seo_2023_CVPR, seo2023flipnerf} employs a range of regularizations derived from depth supervision~\cite{dsnerf, densedepth, wang2023sparsenerf}, appearance~\cite{regnerf,diffusionerf}, semantic consistency~\cite{dietnerf}, visibility~\cite{vipnerf, kwak2023geconerf}, or frequency patterns~\cite{yang2023freenerf}.
Although these techniques have improved the reconstruction quality, undesirable artifacts remain in the synthesized novel views. 
While 3DGS~\cite{3dgs} has improved rendering quality and efficiency using differentiable splatting, it still struggles with the few-view setting, a direction that has not been sufficiently explored.

Diffusion models~\cite{sohl2015deep, ddpm, song2020denoising,rombach2022high,zhang2023adding} have illuminated potential pathways for addressing the few-view NeRF/3DGS reconstruction challenge. 
By lifting the comprehensive visual priors learned from vast 2D datasets into 3D, there have been significant advancements in 3D content generation\footnote{Note that 3D content generation from images differs fundamentally from few-view reconstruction, due to the need to satisfy the multi-view geometry constraint. This work tackles the latter, where the goal is ``reconstruction'' rather than ``generation''.}~\cite{deng2022nerdi, Xu_2022_neuralLift, realfusion, sparsefusion, zero123, chan2023genvs, gu2023nerfdiff, wang2023score}.
Given this backdrop, an intuitive and investigated approach to utilizing 2D diffusion models for sparse view reconstruction is to employ them as a  ``scorer'' evaluating the quality of rendered images and thus a regularizer for NeRF training similar to score distillation sampling (SDS)~\cite{dreamfusion}.
This approach~\cite{diffusionerf,wu2023reconfusion} however necessitates a large diffusion model be inferred at each training step of the radiance field, a computationally intensive process.

In this paper, we propose \textsc{Deceptive-NeRF/3DGS} to efficiently leverage large diffusion models for sparse view reconstruction.
Instead of using diffusion models only as a means to regularize the NeRF/3DGS training process, we produce {\em pseudo-observations} by a diffusion model, directly applying them to the training pipeline as if they were real input views, as shown in \Cref{fig:overall_method}. 
Specifically, we propose a \emph{deceptive diffusion model} that turns coarsely rendered noisy images from few-view reconstructions into high-quality photorealistic pseudo-observations.
To achieve seamless integration of real input views and synthetic pseudo-observations in the training process, we develop a reconstruction uncertainty measure to guide the diffusion model’s generation. 

With a trained deceptive diffusion model, our method alternates between the two steps (\Cref{fig:method}): 1) feeding rendered images into a diffusion model for generating pseudo-observations and 2) optimizing the representation using both the real input images and the newly generated pseudo-observations.
Consequently, this novel approach tackles the issue of sparsity by  ``densifying'' observations by 5 to 10 times, while not demanding excessive time or computation, thanks to the one-time usage of diffusion models.
As a by-product of such densification of input observations, our method acquires an additional feature at no extra cost—rendering novel views with super-resolution. This enables our approach to perform high-quality novel view synthesis even under the stringent conditions of few-view, low-resolution settings.

In summary, our contributions include the following:
\begin{itemize}
    \item We propose a novel approach for few-view reconstruction, agnostic to the underlying scene representation. 
    \item Our approach leverages large diffusion models to generate pseudo-observations and densify input observations fivefold to tenfold. Compared to using diffusion models as a ``scorer'' to regularize the training process, this manner largely reduces the computational cost, with nearly ten times the training speed.
    \item We propose a deceptive diffusion model to generate high-quality photorealistic pseudo-observations preserving scene semantics and view consistency.
    \item Our method can render novel views with super-resolution, surpassing the limitations posed by low-quality inputs.
\end{itemize}

\section{Related Work}
\noindent\textbf{Neural 3D Representations.}
Using continuous 3D fields modeled by MLPs and volumetric rendering, Neural Radiance Fields (NeRFs)~\cite{nerf} have enabled a new and effective approach for novel view synthesis and reconstruction.
Follow-up works have since emerged to enhance NeRFs and expand their applications~\cite{tewari2022advances}, such as
improving rendering quality~\cite{barron2021mip, barron2022mip, barron2023zip}, 
modeling dynamic scenes~\cite{zhang2021stnerf, park2021nerfies, dnerf, tretschk2021non, wang2022fourier, cao2023hexplane, kplanes, attal2023hyperreel, liu2024gear},
acceleration~\cite{yu2021plenoctrees, yu2021plenoxels}, 
and 3D scene editing~\cite{liu2021editing, zhang2021stnerf, wang2021clip, jang2021codenerf,kobayashi2022decomposing, liu2022unsupervised}.
Recent work has shown that replacing the deep MLPs with a feature voxel grid network can significantly improve training and inference speed~\cite{chen2022tensorf, sun2022direct, fridovich2022plenoxels, muller2022instant}. 
More recently, 3D Gaussian Splatting~\cite{3dgs} further improves visual quality, rendering time, and performance.
Despite significant progress, NeRF, 3DGS, and their variants still struggle to reconstruct high-quality 3D models when there are only a limited number of input views.

\noindent\textbf{Few-view Reconstruction.} 
Several studies have been conducted to enhance NeRF when provided with only sparse observations.
Some approaches pre-train their models utilizing prior knowledge from extensive datasets of 3D scenes to generate novel views from the given sparse observations~\cite{yu2020pixelnerf, mvsnerf, SRF, jang2021codenerf,johari2022geonerf}.
On the other hand, more attempts~\cite{seo2023flipnerf, diffusionerf, vipnerf, wang2023sparsenerf, somraj2023simplenerf, roessle2023ganerf} employ a range of regularizations to the NeRF training pipeline.
Among them, \cite{densedepth, dsnerf} use the estimated depth information from Structure-from-Motion (SfM) or depth estimation networks as supplementary supervision.
FreeNeRF~\cite{yang2023freenerf} regularizes the visible frequency range of NeRF's inputs to avoid overfitting. 
\cite{dietnerf, regnerf, diffusionerf} impose regularization on rendered patches from semantic consistency, geometry, and appearance. 
Other attempts include the use of cross-view pixel matching~\cite{sparf}, cross-view feature matching~\cite{chen2023matchnerf,du2023cross}, ray-entropy regularization~\cite{kim2022infonerf}, and visibility priors~\cite{vipnerf,kwak2023geconerf}.
Yet, existing approaches still fall short of the requirement of consumer-level usage and necessitate scene-specific heuristic adjustments.
Our approach achieves improvements over the previously mentioned few-view NeRF techniques and, for the first time, enhances 3DGS performance in the few-view setting.

\noindent\textbf{Diffusion models for 3D.} 
By capitalizing on powerful 2D diffusion models, many works have advanced the frontier of 3D computer vision tasks, such as 3D content generation~\cite{dreamfusion,lin2022magic3d,chen2023single,holodiffusion, realfusion,deng2022nerdi,sparsefusion,gu2023nerfdiff,sparsefusion}.
To achieve this, \cite{zero123} uses a diffusion model trained on synthetic data as geometric priors to synthesize novel views given one single image. 
Closer to our work, DiffusioNeRF~\cite{diffusionerf} and ReconFusion~\cite{wu2023reconfusion} regularize few-view NeRFs with priors from diffusion models. 
Unlike these methods that utilize diffusion models in a 3D setting, our approach does not employ them as a ``scorer'' for regularization. Instead, we use the images generated by the diffusion model as auxiliary pseudo-observations directly for NeRF training, treating them as if they were real input views. As a result, our method avoids inferring the diffusion model at every training step, reducing computational costs and achieving nearly ten times the training speed.

\section{Preliminaries}
\label{sec:preliminary}
\noindent\textbf{Neural Radiance Fields.} 
A radiance field is a continuous function $f$ mapping a 3D coordinate $\mathbf{x}\in\mathbb{R}^3$  and a viewing directional unit vector $\mathbf{d}\in\mathbb{S}^2$ to a volume density  $\sigma\in[0,\infty)$ and RGB values $\mathbf{c}\in[0,1]^3$.
A neural radiance field (NeRF)~\cite{nerf} uses a multi-layer perceptron (MLP) to parameterize this function:
\begin{equation}
    f_\theta:(\mathbf{x},\mathbf{d}) \mapsto (\sigma,\mathbf{c})
\end{equation} where $\theta$ denotes MLP parameters.
While existing NeRF variants employ explicit voxel grids~\cite{yu2021plenoctrees,yu2021plenoxels,chen2022tensorf} instead of MLPs to parameterize this mapping for improved efficiency,
our proposed approach is compatible with both MLP-based NeRFs and voxel grid-based variants.
Rendering each image pixel given a neural radiance field $f_\theta$ involves casting a ray $\mathbf{r}(t)=\mathbf{o}+t\mathbf{d}$ from the camera center $\mathbf{o}$ through the pixel along direction $\mathbf{d}$. The predicted color for the corresponding pixel is computed as:
\begin{equation}
\label{eqn:vanillanerf}
    \hat{\mathbf{C}} = \sum_{k=1}^K \hat{T}(t_k)\alpha(\sigma(t_k)\delta_k)\mathbf{c}(t_k), 
\end{equation}
where $\hat{T}(t_k)=\operatorname{exp}\left(-\sum_{k^{'}=1}^{k-1}\sigma(t_k)\delta(t_k)\right)$, $\alpha \left({x}\right) = 1-\exp(-x)$, and $\delta_p = t_{k+1} - t_k$.
A vanilla NeRF is optimized over a set of input images and their camera poses by minimizing the mean squared error (photometric loss):
\begin{equation}
    \mathcal{L}_\text{pho}=\sum_{\mathbf{r}\in\mathcal{R}}\|\hat{\mathbf{C}}(\mathbf{r)}-\mathbf{C}(\mathbf{r})\|_2^2,
\end{equation}

\noindent\textbf{3D Gaussian Splatting.} 3DGS~\cite{3dgs} represents a 3D scene explicitly through a collection of 3D Gaussians.
Each of the Gaussians is parameterized with a position vector $\boldsymbol{\mu} \in \mathbb{R}^3$ and a covariance matrix $ \mathbf{\Sigma}\in \mathbb{R}^{3\times 3}$. 
Each Gaussian influences a point $\mathbf{x}$ in 3D space following:
\begin{equation}
\label{eq:3dgs}
    G(\mathbf{x})=\frac{1}{\left(2\pi\right)^{3/2}\left|\mathbf{\Sigma}\right|^{1/2}}e^{-\frac{1}{2}{\left(\mathbf{x}-\boldsymbol{\mu}\right)}^T\mathbf{\Sigma}^{-1}\left(\mathbf{x}-\boldsymbol{\mu}\right)}.
\end{equation}
Each Gaussian stores an opacity logit $ o \in \mathbb{R}$ and the appearance feature is represented by spherical harmonic (SH) coefficients.
To render the RGB color of a pixel, one orders all the Gaussians that contribute to it and blends the ordered Gaussians using:
\begin{equation}
\label{eq:render}
          \hat{\mathbf{C}} = \sum_{i=1}^n\ \mathbf{c}_i \alpha_i  \prod_{j=1}^{i-1}(1-\alpha_j),
\end{equation}
where $\mathbf{c}_i$ is the color computed from the SH coefficients. $\alpha_i$ is given by evaluating a 2D Gaussian multiplied by the opacity logit $o$.

\section{Method}
Given only $W$ observations of a scene, i.e., input images $\mathcal{I} = \{I_1,I_2,\dots,I_W\}$ with their calibrated camera poses, our approach enhances the NeRF reconstruction through employing diffusion models to ``densify'' the inputs, as illustrated in \Cref{fig:method}. 
In \Cref{sec:densify}, we describe our pipeline, which continuously densifies the input observations through synthetic pseudo-observations generated by diffusion models during the NeRF training process.
In \Cref{sec:uncertainty}, we propose an uncertainty measure that accompanies rendered RGB images to enforce 3D consistency on the generated pseudo-observations.
In \Cref{sec:ldm}, we introduce our proposed deceptive diffusion model, that can generate high-quality photorealistic pseudo-observations conditioned on coarse-rendered images and uncertainty maps.

\subsection{Pipeline Overview}
\label{sec:densify}
We start by training the NeRF/3DGS representation using the few-view input images.
After a set number of epochs, we obtain a coarse reconstruction, 
from which we can generate RGB images, depth predictions, as well as uncertainty maps at novel views. 
To synthesize pseudo-observation, we randomly sample novel views within a bounding box defined by the outermost input views.
At novel viewpoint $A$, we render RGB image $\hat{I}_A$ and depth map $D_A$ and further compute an uncertainty map $U_A$. We defer describing the process of obtaining $D_A$ and $U_A$ until \Cref{sec:uncertainty}.
We generate a pseudo-observation at viewpoint $A$, denoted as $P_A$, using our proposed deceptive diffusion model.
Thanks to the natural image prior from the latent diffusion model and multi-view uncertainty measures, the pseudo-observations could eliminate the artifacts in $\hat{I}_A$.
After accumulating a sufficient number of pseudo-observations, we continue training the NeRF representation, using the combination of both the original input images (real) and pseudo-observations (synthesized). 
We discard 20\% of the pseudo-observations with the lowest perceptual similarity to input images, quantified through the LPIPS metric.
This helps avoid generating content that is far from the original inputs.  

We iterate the above processes, each time sampling new viewpoints to render images and generating corresponding pseudo-observations, enhancing the training image set until we accumulate more than a sufficient number of observations. 
In other words, we alternate between the two steps to get a final reconstruction: 1) feeding NeRF-rendered images into a diffusion model for generating pseudo-observations and 2) optimizing the NeRF representation using both the real input images and the newly generated pseudo-observations.

\subsection{Uncertainty Measure for View-Consistency}
\label{sec:uncertainty}
To ensure consistency between the synthesized pseudo-observations and real input views, we further guide the diffusion model with an uncertainty measure.
While there are approaches to quantify the uncertainty from feature space perspectives through ensemble learning~\cite{sunderhauf2023density}, variational inference~\cite{shen2022conditional} or spatial perturbations~\cite{goli2023bayes}, we instead acquire uncertainty maps through correspondence matching for rendered RGB images to explicitly enforce view consistency.
Compared to \cite{sunderhauf2023density, shen2022conditional, goli2023bayes},  our proposed measure of uncertainty does not involve modifying the underlying representation and only requires reprojecting real input observations to novel views, thus minimizing additional computation while enforcing epipolar constraints.
Given a rendered RGB image at a novel view, we leverage its associated rendered depth map to warp the image at its nearest input view to the unobserved viewpoint.
Similar to \Cref{eqn:vanillanerf}, we compute the depth value for each ray as a weighted composition of distances traveled from the origin as:
\begin{equation}
    \hat{\mathbf{D}} = \sum_{k=1}^K \hat{T}(t_k)\alpha(\sigma(t_k)\delta_k).
\end{equation}
At an arbitrary novel view $A$, after getting the rendered image $\hat{I}_A$,  we use its associated rendered depth map $D_A$ to warp the ground truth image $I_\text{B}$ towards $A$, where $B$ is its nearest input viewpoint.
This results in a warped image 
\begin{equation}
    I_{B\rightarrow A} = \psi(I_\text{B}; D_A, R_{B\rightarrow A}),
\end{equation} where $R_{B\rightarrow A}$ is the viewpoint difference between $A$ and $B$. 
More specifically, each pixel location $p$ in $\hat{I}_A$ is transformed to $q$ in $I_B$ as:
\begin{equation}
q = KR_{B\rightarrow A}{D}_A(p)K^{-1}p,
\end{equation}
where $K$ is the camera parameter matrix. 
We define the uncertainty map for the viewpoint $A$ as the squares of the distance between $I_{B \rightarrow A}$ and $\hat{I}_A$:
\begin{equation}
    U_A = (I_{B \rightarrow A} - \hat{I}_A)^2.
\end{equation}
Our proposed deceptive diffusion model will condition on $U_A$ to generate pseudo-observations with view-consistency guidance.

\subsection{Deceptive Diffusion Model}
\label{sec:ldm}
We propose a 2D diffusion model $\operatorname{G}$ that conditions on a NeRF-rendered RGB image $\hat{I}_A$ along with its corresponding depth prediction $D_A$ and uncertainty map $U_A$ to generate a refined natural image as the pseudo-observation $P_A$:
\begin{equation}
   P_A = \operatorname{G}(\hat{I}_A, D_A, U_A),
\end{equation}
where $\operatorname{G}$ in essential rectifies renderings from NeRF/3DGS and is thus termed the deceptive diffusion model.

Our approach capitalizes on latent diffusion models~\cite{rombach2022high}, which leverages natural image priors derived from internet-scale data to help ameliorate unnaturalness caused by sparse input observations.
Artifacts generated by NeRFs often float in empty space and are therefore highly conspicuous in depth prediction. Conditioning the diffusion model also on the depth prediction takes advantage of this. On the other hand, the uncertainty map introduced in \Cref{sec:uncertainty} provides cross-view constraints for the generation of pseudo-observations.

To this end, given a dataset of quartets $\left\{\left(I_\text{fine},I_\text{coarse},D,U\right)\right\}$, we fine-tune a pre-trained diffusion model, consisting of a latent diffusion architecture with an encoder $\mathcal{E}$, a denoiser U-Net $\epsilon_\theta$, and a decoder $\mathcal{D}$.
We solve for the following objective to fine-tune the model:
\begin{equation}
    \min_{\theta}\mathbb{E}_{z\sim \mathcal{E},t,\epsilon\sim\mathcal{N}(0,1)}\|\epsilon-\epsilon_{\theta}(z_t,t,c(I_\text{coarse},D,U,s))\|_2^2,
\end{equation}
where the diffusion time step $t\sim [1,1000]$ and $c(I_\text{coarse},D,U,s)$ is the embedding of the coarse RGB image, depth estimation, uncertainty map, and a text embedding $s$ of the coarse image. 
To enable diffusion models to learn such specific input conditions without disrupting their prior for natural images, we leverage ControlNet~\cite{zhang2023adding} to efficiently implement the training paradigm discussed below while preserving the production-ready weights of pre-trained 2D diffusion models. 

\noindent\textbf{Text embedding.} To derive text embedding $s$ from the coarse rendered image, we first generate a text prompt $s_0$ using a pre-trained image captioning network. 
While image captioning reliably provides descriptive textual representations for most coarse images, its efficacy can diminish for images of lower quality or those with pronounced artifacts.
To counteract this, we adopt the textual inversion~\cite{gal2022image}. 
We optimize a shared latent text embedding $s_*$ shared by all the input observations and coarse images.
By concatenating the embeddings we formulate a composite feature $s=[s_0,s_{*}]$ that encapsulates both the semantic and visual attributes of the input image. This combined strategy not only ameliorates the shortcomings of image captioning but also ensures the stylistic congruence of the generated pseudo-observations with the input images.

\noindent\textbf{Data augmentation.} 
To train the deceptive diffusion model to generate pseudo-observations from the same viewpoint's rendered RGB image, depth map, and uncertainty map, we need to construct a dataset of quartets $\left(I_\text{fine}, I_\text{coarse}, D, U\right)$. 
$I_\text{fine}$ is a fine image with minimal artifacts, closely resembling a real image or a perfectly rendered NeRF image. 
$I_\text{coarse}$ is a coarse image from the same viewpoint, likely presenting obvious artifacts, while $D$ and $U$ represent the associated depth map and uncertainty map, respectively.
We can construct such a dataset by training two versions of NeRF for the same scene: a fine version of NeRF trained on all images and a coarse version of NeRF trained on a small subset of input images. 
By rendering from the same viewpoint, such a coarse-fine NeRF duo can render paired training data samples, as well as depth and uncertainty maps.
However, due to limited computational resources, we cannot afford to conduct NeRF duos training across a plethora of scenarios.
Therefore, we introduce a data augmentation paradigm to mitigate the computational cost associated with preparing training data.
Rather than exclusively relying on image pairs derived from NeRF duos, we exploit a more straightforward data source during the initial phase of training. We add random Gaussian noise to RGB images, utilizing these noisy images and accompanying depth maps as training inputs, while retaining the original RGB images as the training objectives.
In this manner, we can readily obtain training samples by simply pairing RGB-depth data.
Despite the gap between the data obtained in this manner and the data previously discussed from NeRF duos, we find that they effectively enable diffusion models to better utilize depth maps to understand imperfect RGB images.
(When training with such additional data, we do not feed uncertainty maps into the diffusion model.)

\section{Experimental Results}
In this section, we assess the performance of our proposed Deceptive-NeRF for the task of few-shot novel view synthesis across a range of challenging datasets, comparing it with state-of-the-art methods. 
We introduce the setup of our experiments in \Cref{sec:setup}.
To demonstrate our method's capability for super-resolution on novel views, we showcase experimental results under the challenging setting of few views and low resolution in \Cref{sec:sr}.
To elucidate the effectiveness of our design, we conduct a series of evaluations of our approach, including analyses of the input densification scale and generated pseudo-observations,  as well as an ablation study of other designs and components in \Cref{sec:ablation}.

\subsection{Experimental Setup.}
\label{sec:setup}
\noindent\textbf{Implementation Details.}
We train our deceptive diffusion model on a mixture of Hypersim~\cite{roberts2021hypersim} and CO3D~\cite{co3d}.
For training data augmentation, we corrupt 60,000 images from Hypersim~\cite{roberts2021hypersim} by adding additive Gaussian noise with a standard deviation of $0.3$. 
We use these noisy images and their depth maps as training input and the original images as training targets, without running NeRF reconstruction. 
For the major section of the training dataset for our deceptive diffusion model, we train coarse and fine NeRF duos for the same scenes, where coarse NeRFs are trained only with one-fifth of input images.
\begin{figure}[t]
\begin{center}
    \includegraphics[width=\linewidth]{ 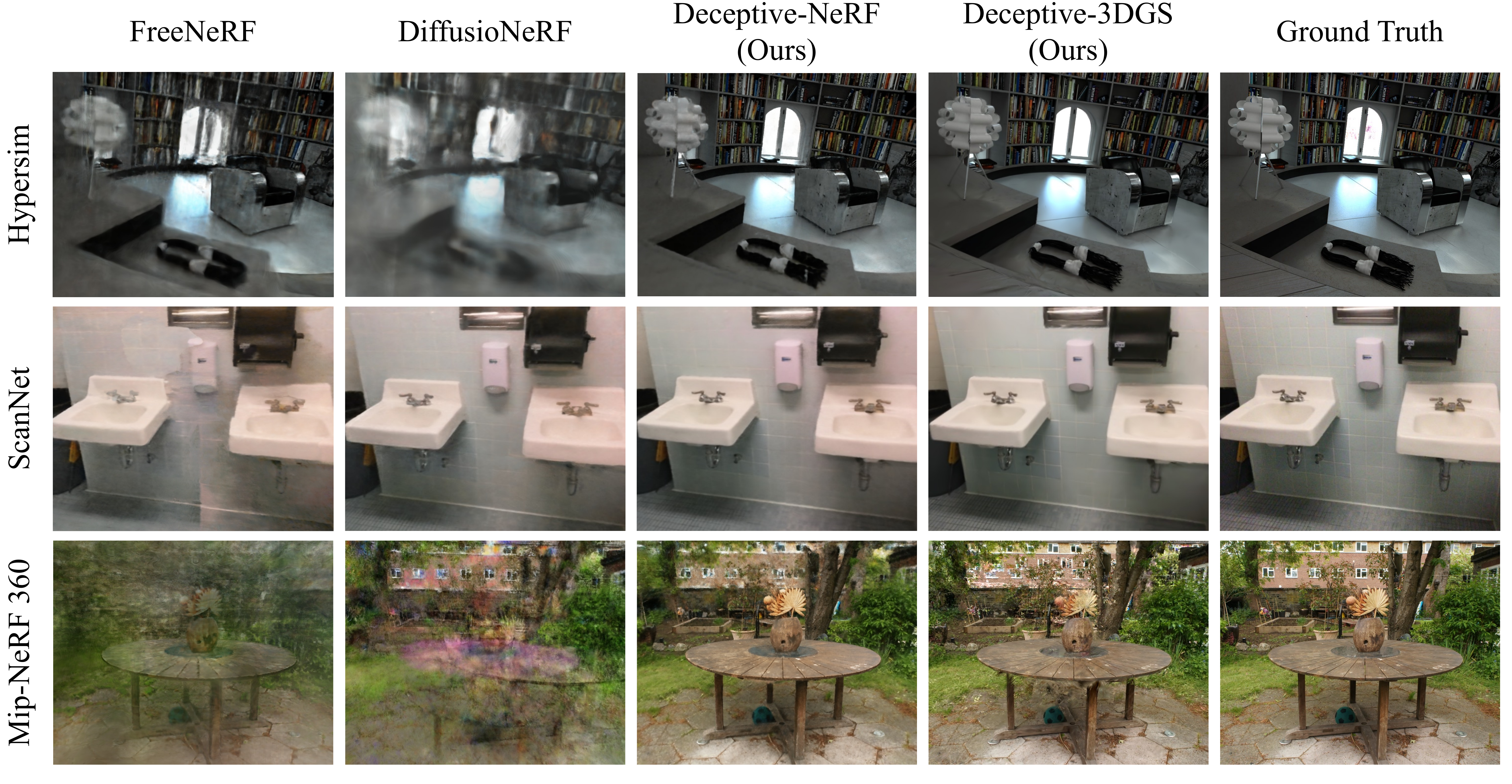}
\end{center}
\vspace{-5mm}
\caption{\textbf{Qualitative comparisons of few view reconstruction.}  Scenes are from the Hypersim~\cite{roberts2021hypersim}, Scannet~\cite{scannet} and mip-NeRF 360~\cite{barron2022mipnerf360} datasets, with 10 input views.
}
\vspace{-3mm}
\label{fig:comparison}
\end{figure}
\begin{table}[t]
    \centering
    \caption{\textbf{Quantitative comparison on Hypersim.} 
    \vspace{-3mm}
    \colorbox{topcolor}{best} \colorbox{secondcolor}{second-best} \colorbox{thirdcolor}{third-best}}
    \label{table:hypersim}
    \fontsize{8.1}{8}\selectfont
    \begin{tabular}{
        l
        |S[table-format=2.2]
        S[table-format=2.2]
        S[table-format=2.2]
        |S[table-format=1.3]
        S[table-format=1.3]
        S[table-format=1.3]
        |S[table-format=1.3]
        S[table-format=1.3]
        S[table-format=1.3]
    }
        \toprule
        & \multicolumn{3}{c|}{PSNR$(\uparrow)$} & \multicolumn{3}{c|}{SSIM$(\uparrow)$} & \multicolumn{3}{c}{LPIPS$(\downarrow)$} \\
        Method & {5-view} & {10-view} & {20-view} & {5-view} & {10-view} & {20-view} & {5-view} & {10-view} & {20-view} \\
        \midrule
        PixelNeRF & 7.76 & 8.31 & 10.90 & 0.221 & 0.380 & 0.374 & 0.542 & 0.571 & 0.503 \\
        MVSNeRF & 11.58 & 12.00 & 14.42 & 0.271 & 0.274 & 0.315 & 0.563 & 0.512 & 0.457 \\
        DS-NeRF & 13.79 & 13.66 & 18.80 & 0.388 & 0.431 & 0.488 & 0.515 & 0.511 & 0.481  \\
        DietNeRF & 13.01 & 13.51 & 18.62 & 0.417 & 0.479 & 0.481 & 0.541 & 0.527 & 0.472 \\
        RegNeRF & 15.65 & \cellcolor{thirdcolor}18.59 & 19.26 & 0.491 & 0.501 & 0.519 & 0.516 & 0.451 & 0.362 \\
        DiffusioNeRF & 16.40 & 17.22 & 19.88 & 0.451 & 0.470 & 0.656 & 0.432 & 0.404 & 0.416 \\
        FlipNeRF & 15.43 & 17.47 & 19.36 & 0.456 & 0.569 & 0.585 & \cellcolor{thirdcolor}0.350 & 0.415 & 0.312 \\
        FreeNeRF & \cellcolor{thirdcolor}17.20 & 18.06 & \cellcolor{thirdcolor}20.20 & \cellcolor{thirdcolor}0.599 & \cellcolor{thirdcolor}0.671 & \cellcolor{thirdcolor}0.706 & 0.431 & \cellcolor{thirdcolor}0.286 & \cellcolor{thirdcolor}0.237 \\
        Deceptive-NeRF (Ours) & \cellcolor{secondcolor}18.91 & \cellcolor{secondcolor}19.88 & \cellcolor{secondcolor}21.23 & \cellcolor{secondcolor}0.652 & \cellcolor{secondcolor}0.726 & \cellcolor{secondcolor}0.761 & \cellcolor{secondcolor}0.322 & \cellcolor{secondcolor}0.270 & \cellcolor{topcolor}0.222 \\
        Deceptive-3DGS (Ours) & \cellcolor{topcolor}19.31 & \cellcolor{topcolor}21.45 & \cellcolor{topcolor}21.61 & \cellcolor{topcolor}0.728 & \cellcolor{topcolor}0.774 & \cellcolor{topcolor}0.788 & \cellcolor{topcolor}0.265 & \cellcolor{topcolor}0.241 & \cellcolor{secondcolor}0.228 \\
        \bottomrule
    \end{tabular}
    \vspace{-3mm}
\end{table}
Coarse NeRFs render RGB images and depth maps as training inputs while fine NeRFs render fine RGB images from the same viewpoints as training targets.
We use a total of 200 scenes and approximately 20,000 images for the data generation of this stage, with Hypersim~\cite{roberts2021hypersim} and CO3D~\cite{co3d} each contributing half of them.
We fine-tune a pre-trained Stable Diffusion model with ControlNet~\cite{zhang2023adding} into our deceptive diffusion model.
We set five (3 for coarse RGB, 1 for depth, and 1 for uncertainty) control map channels to match the inputs and use all default parameters for the fine-tuning task. 
We randomly crop all images to a resolution of 512 $\times$ 512.

\noindent\textbf{Datasets.} To validate the performance and generalizability of our method for sparse view reconstruction, we employed a series of datasets with distinct characteristics, including Hypersim~\cite{roberts2021hypersim}, a synthetic indoor dataset (from which we avoided scenes already used in training the deceptive diffusion model), ScanNet~\cite{scannet}, a real-world indoor dataset, LLFF~\cite{mildenhall2019local}, a real-world facing-forward dataset, and Mip-NeRF 360~\cite{barron2022mipnerf360}, a real-world 360-degree outdoor dataset. 

\noindent\textbf{Evaluation Metrics.} We evaluate using the following standard metrics: (i) Peak Signal-to-Noise Ratio (PSNR), (ii) Structural Similarity Index Measure (SSIM)~\cite{wang2004image} and (iii) Learned Perceptual Image Patch Similarity (LPIPS)~\cite{zhang2018unreasonable}, by comparing the reconstructed frames against ground truth images. These metrics are computed on the held-out view and averaged across all frames.

\noindent\textbf{Baselines. }We compare our method with several methods within a similar scope.
PixelNeRF~\cite{yu2020pixelnerf}, MVSNeRF~\cite{mvsnerf}, and SRF~\cite{SRF} are representative pre-trained methods, exploiting the DTU and LLFF datasets for pre-training.
We also compare our approach against diverse regularization approaches, including DS-NeRF~\cite{dsnerf}, DietNeRF~\cite{dietnerf}, RegNeRF~\cite{regnerf}, DiffusioNeRF~\cite{diffusionerf}, FlipNeRF~\cite{seo2023flipnerf}, and FreeNeRF~\cite{yang2023freenerf}.

\subsection{Comparisons of Few View Reconstruction}
\label{sec:comp_recon}
We show the qualitative results in \Cref{fig:comparison} and report the quantitative comparisons in \Cref{table:hypersim} and \Cref{table:LLFF}, with more results showcased in the supplementary material. 
On challenging datasets such as those featuring indoor scenes facing outwards or 360-degree outdoor scenes, existing state-of-the-art methods can easily generate discernible artifacts.  By utilizing pseudo-observations to densify inputs, our method avoids such artifacts produced by overfitting.
Both Deceptive-NeRF and Deceptive-3DGS surpass the baseline methods across all metrics, demonstrating high-quality reconstruction under various challenging scenes and with different numbers of inputs.

\begin{table}[t]
    \centering
    \caption{\textbf{Quantitative comparison on LLFF.}   \colorbox{topcolor}{best} \colorbox{secondcolor}{second-best} \colorbox{thirdcolor}{third-best}}
    \label{table:LLFF}
    \fontsize{8.1}{8}\selectfont
    \begin{tabular}{l
        |S[table-format=2.2]
        S[table-format=2.2]
        S[table-format=2.2]
        |S[table-format=1.3]
        S[table-format=1.3]
        S[table-format=1.3]
        |S[table-format=1.3]
        S[table-format=1.3]
        S[table-format=1.3]}
        \toprule
        & \multicolumn{3}{c|}{PSNR$(\uparrow)$} & \multicolumn{3}{c|}{SSIM$(\uparrow)$} & \multicolumn{3}{c}{LPIPS$(\downarrow)$} \\
        Method & {3-view} & {6-view} & {9-view} & {3-view} & {6-view} & {9-view} & {3-view} & {6-view} & {9-view} \\
        \midrule
        SRF & 17.07 & 16.75 & 17.39 & 0.436 & 0.438 & 0.465 & 0.529 & 0.521 & 0.503 \\
        PixelNeRF & 16.17 & 17.03 & 18.92 & 0.438 & 0.473 & 0.535 & 0.512 & 0.477 & 0.430 \\
        MVSNeRF & 17.88 & 19.99 & 20.47 & 0.584 & 0.660 & 0.695 & 0.327 & 0.264 & 0.244 \\
        DietNeRF & 14.94 & 21.75 & 24.28 & 0.370 & 0.717 & 0.801 & 0.496 & 0.248 & 0.183 \\
        RegNeRF & 18.84 & 23.22 & 24.88 & 0.573 & 0.770 & 0.826 & 0.345 & 0.203 & 0.159\\
        FreeNeRF & \cellcolor{thirdcolor}19.63 & \cellcolor{thirdcolor}23.73 & \cellcolor{thirdcolor}25.13 & \cellcolor{thirdcolor}0.612 & \cellcolor{thirdcolor}0.779 & \cellcolor{thirdcolor}0.827 & \cellcolor{thirdcolor}0.308 & \cellcolor{thirdcolor}0.195 & \cellcolor{thirdcolor}0.160  \\
        Deceptive-NeRF (Ours)  & \cellcolor{secondcolor}19.87 &\cellcolor{secondcolor} 23.92 & \cellcolor{topcolor}25.34 & \cellcolor{topcolor}0.688 & \cellcolor{topcolor}0.810 & \cellcolor{secondcolor}0.834 & \cellcolor{secondcolor}0.304 & \cellcolor{secondcolor}0.190 & \cellcolor{secondcolor}0.155 \\
        Deceptive-3DGS (Ours)  & \cellcolor{topcolor}19.95&\cellcolor{topcolor} 24.15 & \cellcolor{secondcolor}25.30 & \cellcolor{secondcolor}0.665 & \cellcolor{secondcolor}0.795 & \cellcolor{topcolor}0.838 & \cellcolor{topcolor}0.295 & \cellcolor{topcolor}0.177 & \cellcolor{topcolor}0.150 \\
        \bottomrule
    \end{tabular}
\end{table}

\begin{figure}[t]
\begin{center}
    \includegraphics[width=\linewidth]{ 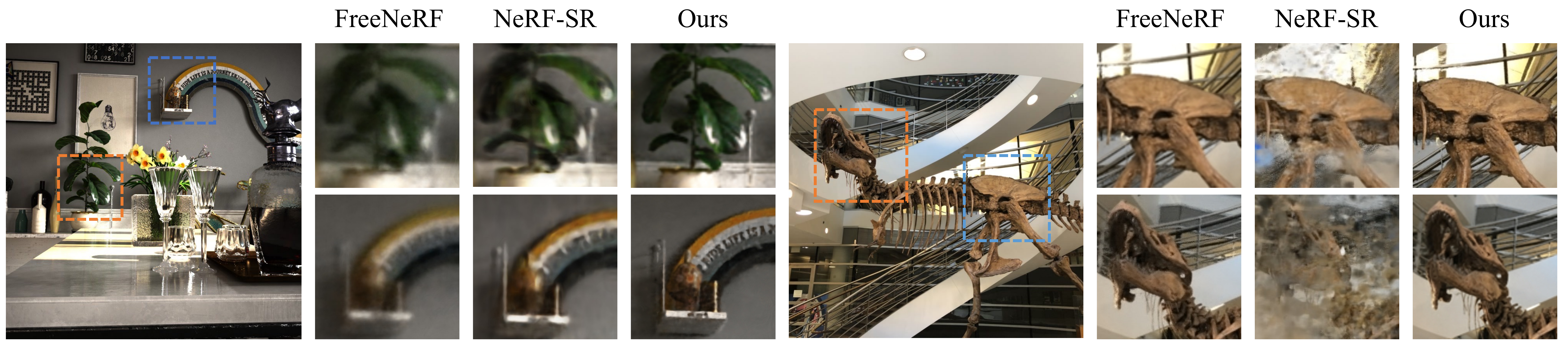}
\end{center}
\vspace{-0.5cm}
\caption{\textbf{Super-Resolution Capabilities.} Our Deceptive-NeRF, when applied to the Hypersim and LLFF datasets with input images downsampled by a factor of 4, demonstrates exceptional super-resolution performance, capable of recovering intricate details.
}
\vspace{-0.25cm}
\label{fig:sr}
\end{figure}

\subsection{Comparisons of Novel View Super-Resolution}
\label{sec:sr}
Due to the generation of high-quality pseudo-observations, our method is capable of super-resolving synthesized novel views under low-resolution and sparse-view conditions.
We train Deceptive-NeRF on the Hypersim dataset~\cite{roberts2021hypersim} with 10 input views and the LLFF dataset~\cite{mildenhall2019local} with 6 input views. All input images were downsampled by a factor of 4 and rendered in multiple views at the original resolution.
As shown in \Cref{fig:sr}, our results can achieve high-quality super-resolution in novel views and recover fine details such as the glossiness of plant leaves and the texture of dinosaur skeletons.
We compared Deceptive-NeRF with the few-view reconstruction approach FreeNeRF~\cite{yang2023freenerf} and the NeRF super-resolution method NeRF-SR~\cite{wang2022nerf}. FreeNeRF, due to not incorporating 2D diffusion priors, fails to restore complex object details, while NeRF-SR suffers from performance degradation due to sparse observations and creates artifacts.

\subsection{Evaluations}
\label{sec:ablation}

\begin{figure}[t]
\begin{center}
    \includegraphics[width=\linewidth]{ 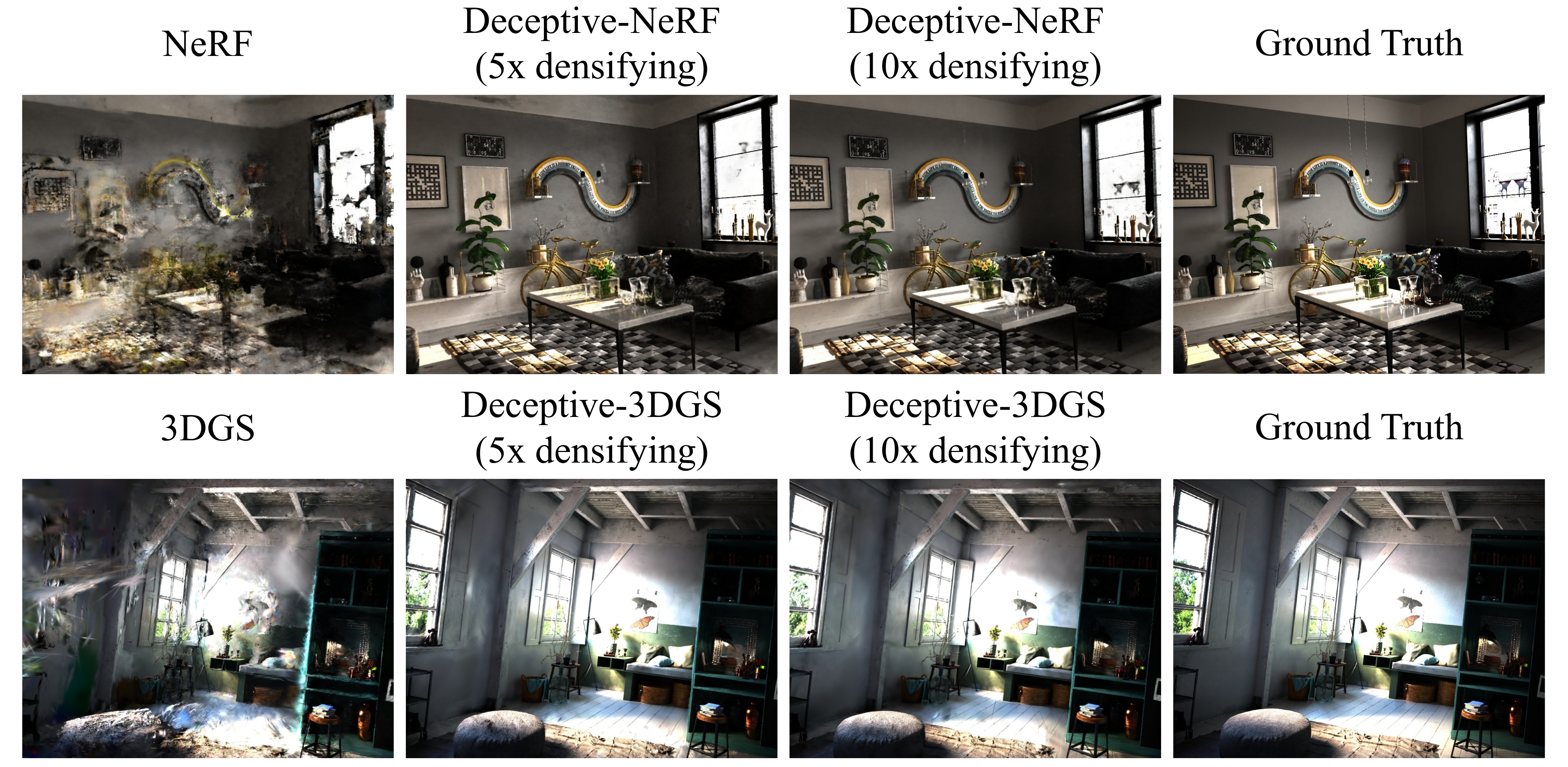}
\end{center}
\vspace{-0.2in}
\caption{\textbf{Evaluation of Densification scales.} We run our method with observations densified by varying scales on Hypersim~\cite{roberts2021hypersim} with 10 input views. Vanilla NeRF or 3DGS results in severe artifacts; fivefold densification reduces them but lacks detail restoration, whereas tenfold densification achieves high-quality reconstructions.
}
\vspace{-0.2in}
\label{fig:densification}
\end{figure}

\textbf{Densification scales.} To study the impact of the densification scale on the results, we conducted experiments with observations densified by varying scales. We carry out these experiments using the Hypersim dataset~\cite{roberts2021hypersim} with 10 input views. As illustrated in \Cref{fig:densification}, not performing densification (vanilla NeRF or 3DGS) leads to severe artifacts while densifying fivefold significantly reduces artifacts but still falls short of high-quality detail restoration. Densifying tenfold, however, results in high-quality reconstructions. Please refer to the appendix for related quantitative results.

\begin{figure}[t]
\begin{center}
    \includegraphics[width=1.0\linewidth]{ 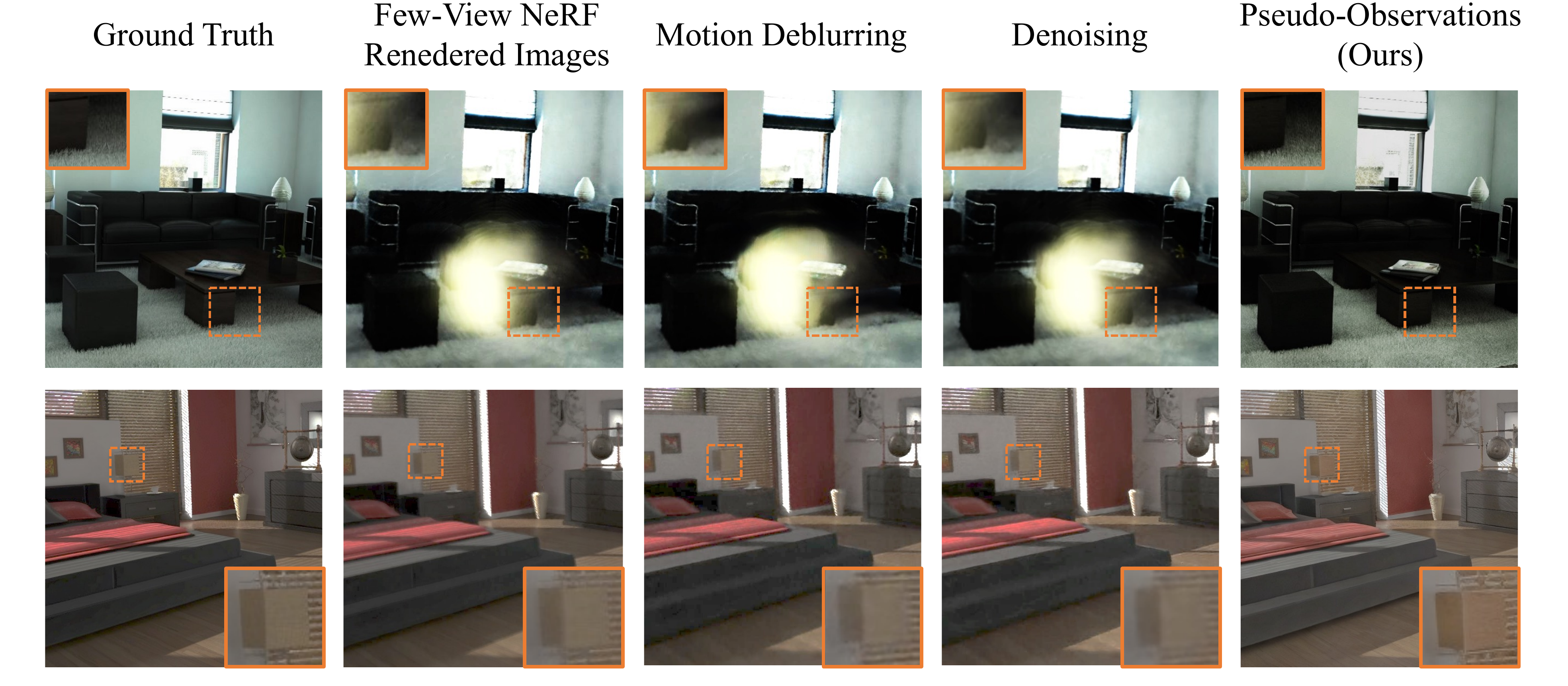}
\end{center}
\vspace{-0.2in}
\caption{\textbf{Visualization of pseudo-observations.} Pseudo-observations synthesized by our deceptive diffusion model remove floating artifacts and blurriness caused by sparse observation in coarse NeRFs, which cannot be achieved by image restoration models.
}\vspace{-0.1in}
\label{fig:pseudo}
\end{figure}

\noindent\textbf{Pseudo-observations.} To better understand our generated pseudo-observations and validate the effectiveness of our deceptive diffusion model, we compare them with images processed using an image restoration model~\cite{zamir2022restormer}.
\Cref{fig:pseudo} shows pseudo-observations synthesized by our deceptive diffusion model effectively remove floating artifacts caused by sparse observation in coarse NeRFs and also mitigate blurriness as well. In contrast, image restoration models (motion deblurring and denoising) fail to convert coarse NeRF images into reasonable pseudo-observations. This is primarily because they are designed and trained for specific image restoration tasks, not for generating pseudo-observations.

\begin{figure}[t]
\begin{center}
    \includegraphics[width=1.0\linewidth]{ 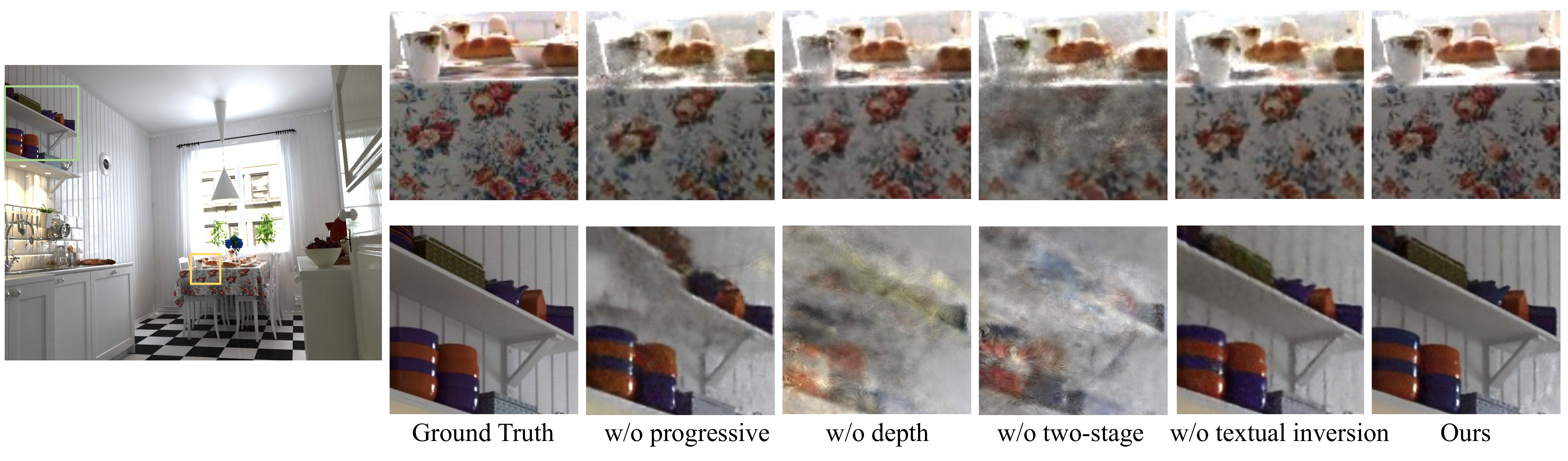}
\end{center}
\vspace{-0.2in}
\caption{\textbf{Qualitative ablation studies.} Our full model synthesizes novel views with the highest quality, while variants tend to produce floating artifacts and blurry details.
}
\vspace{-0.1in}
\label{fig:ablation}
\end{figure}

\noindent\textbf{Ablation studies.} We conduct ablation studies on the following design choices using the Hypersim dataset~\cite{roberts2021hypersim} under the 20-view setting:
\begin{enumerate}
    \item \textbf{Depth Conditioning.} Our deceptive diffusion model generates pseudo-observations conditioned on rendered depth maps. To gauge the significance of this choice, we train a variant that solely conditions on raw RGB images for generating pseudo-observations.
    \item \textbf{Data Augmentation.} We evaluate the impact of our data augmentation procedure when training our deceptive diffusion model. Specifically, we train the model without the initial stage and rely solely on coarse-fine NeRF pairs to generate training samples.
    \item \textbf{Text Embedding.} Our approach 
    integrates both image captioning and textual inversion to address severely artifacted images while ensuring stylistic consistency. We test two variants of our model, one without image captioning and the other without textual inversion.
\end{enumerate}
As illustrated in \Cref{fig:ablation} and \Cref{tab:ablation}, our complete model synthesizes the most photorealistic novel views and outperforms other methods in all quantitative metrics.

\begin{table}[t]
\centering
\caption{\textbf{Quantitative ablation studies.} We ablate our design choices on the Hypersim dataset~\cite{roberts2021hypersim} with 20 input views. \colorbox{topcolor}{best} \colorbox{secondcolor}{second-best} \colorbox{thirdcolor}{third-best}}

\label{tab:ablation}
\fontsize{9}{8}\selectfont
\begin{tabular}{ccccc|S[table-format=2.2]|S[table-format=1.3]|S[table-format=1.3]}
\toprule
Progressive & Depth & Two-stage & Captioning & Inversion& {PSNR $(\uparrow)$} & {SSIM $(\uparrow)$} & {LPIPS $(\downarrow)$} \\
\midrule
 & \checkmark&\checkmark & \checkmark & \checkmark &19.90 & 0.555 & 0.358 \\
\checkmark & & \checkmark & \checkmark & \checkmark & 18.79 & 0.489 & 0.352\\
\checkmark & \checkmark & & \checkmark & \checkmark &  20.49 & 0.619 & 0.290 \\
\checkmark & \checkmark & \checkmark &  &\checkmark  & 21.59 \cellcolor{secondcolor} & 0.758 \cellcolor{secondcolor} & 0.236 \cellcolor{secondcolor} \\
\checkmark & \checkmark & \checkmark & \checkmark &  & 20.58 \cellcolor{thirdcolor} & 0.744 \cellcolor{thirdcolor} & 0.239 \cellcolor{thirdcolor} \\
\checkmark & \checkmark & \checkmark & \checkmark & \checkmark & 22.41 \cellcolor{topcolor} & 0.812 \cellcolor{topcolor}& 0.202 \cellcolor{topcolor}\\
\bottomrule
\end{tabular}
\vspace{-0.1in}
\end{table}

\section{Conclusion} 
We introduce Deceptive-NeRF and Deceptive-3DGS, which utilize 2D diffusion models to generate plausible and consistent pseudo-observations, thereby enhancing sparse view reconstruction by densifying input observations. This fundamentally differs from recent methods that employ 2D diffusion priors to introduce regularization to NeRF training for sparse inputs, by significantly reducing computational costs. Our proposed deceptive diffusion model, guided by an uncertainty measure, can generate high-quality, photorealistic pseudo-observations. Experimental evidence demonstrates that our method can generate pseudo-observations of high quality for synthesizing novel views, surpassing current state-of-the-art approaches. Moreover, our method is capable of rendering novel views with super-resolution, effectively overcoming the challenges posed by low-quality inputs.

\noindent\textbf{Acknowledgements} This work was supported in part by Dartmouth A$\&$S Startup fund.

\bibliographystyle{splncs04}
\bibliography{main}

\appendix

\renewcommand\thefigure{\Alph{figure}} 
\renewcommand\thetable{\Alph{table}}   
\setcounter{figure}{0} 
\setcounter{table}{0} 

\newpage
\section*{Appendix}

\section{Additional Comparisons}
\label{sec:add_comp}
\begin{figure}[t]
\vspace{-0.5cm}
\begin{center}
    \includegraphics[width=\linewidth]{ 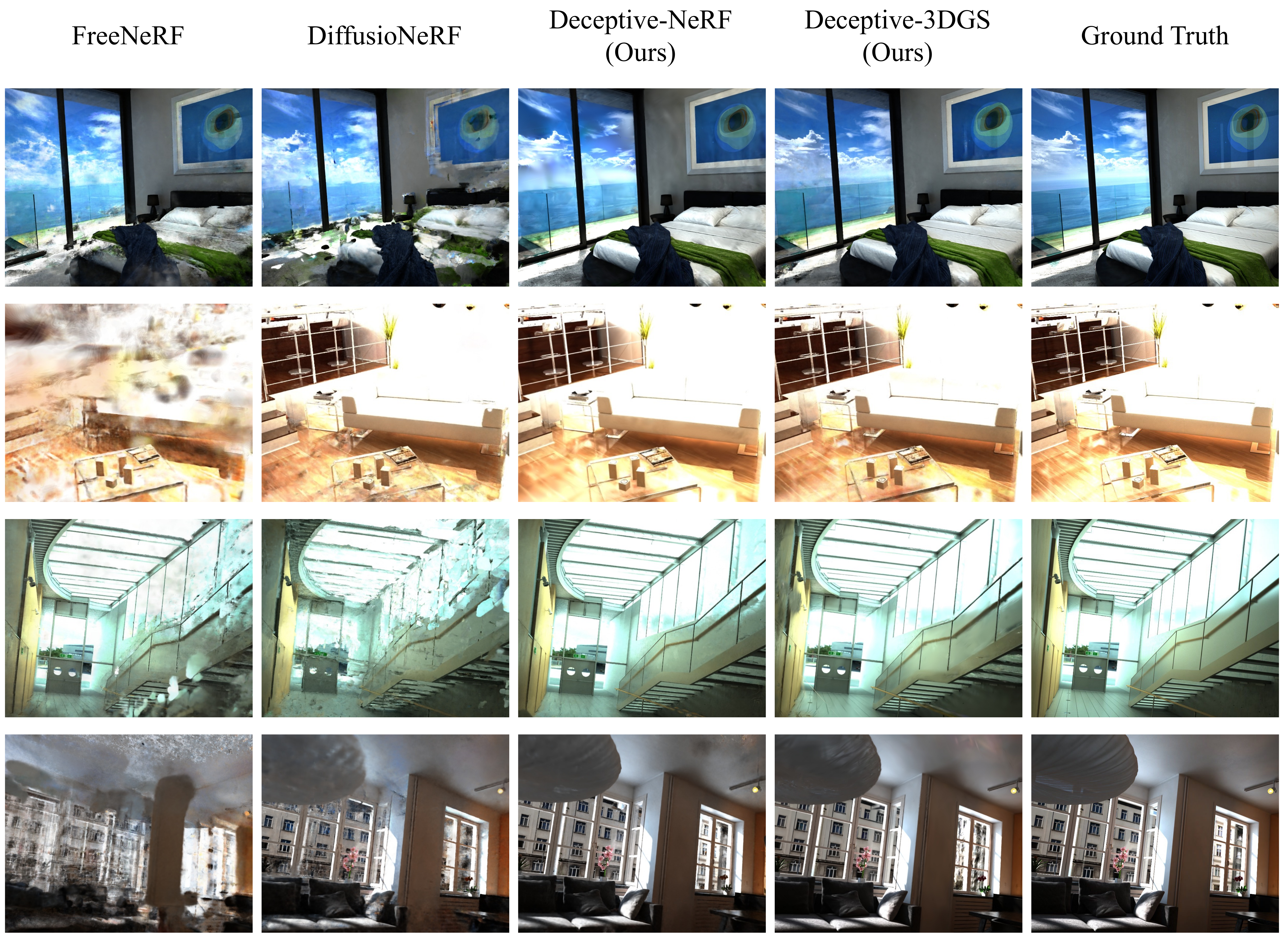}
\end{center}
\vspace{-0.5cm}
\caption{\textbf{Qualitative comparisons of few-view reconstruction on Hypersim dataset~\cite{roberts2021hypersim}.} For each scene, we reconstruct with 10 input views. Novel views synthesized by our proposed Deceptive-NeRF and Deceptive-3DGS do not exhibit floating artifacts. They offer better restoration of distant scenes (such as the sky and buildings viewed seen through the windows).}
\label{fig:hypersim}
\vspace{-0.5cm}
\end{figure}

\noindent\textbf{Sparse-view reconstruction.} To further evaluate our proposed Deceptive-NeRF and Deceptive-3DGS in few-view reconstruction tasks, we compared our method against baseline approaches FreeNeRF~\cite{yang2023freenerf} and DiffusioNeRF~\cite{diffusionerf} in novel view synthesis on mip-NeRF 360~\cite{barron2022mipnerf360} and Hypersim~\cite{roberts2021hypersim} datasets, with 10 input views. 
\begin{figure}[h]
\vspace{-0.5cm}
\begin{center}
    \includegraphics[width=\linewidth]{ 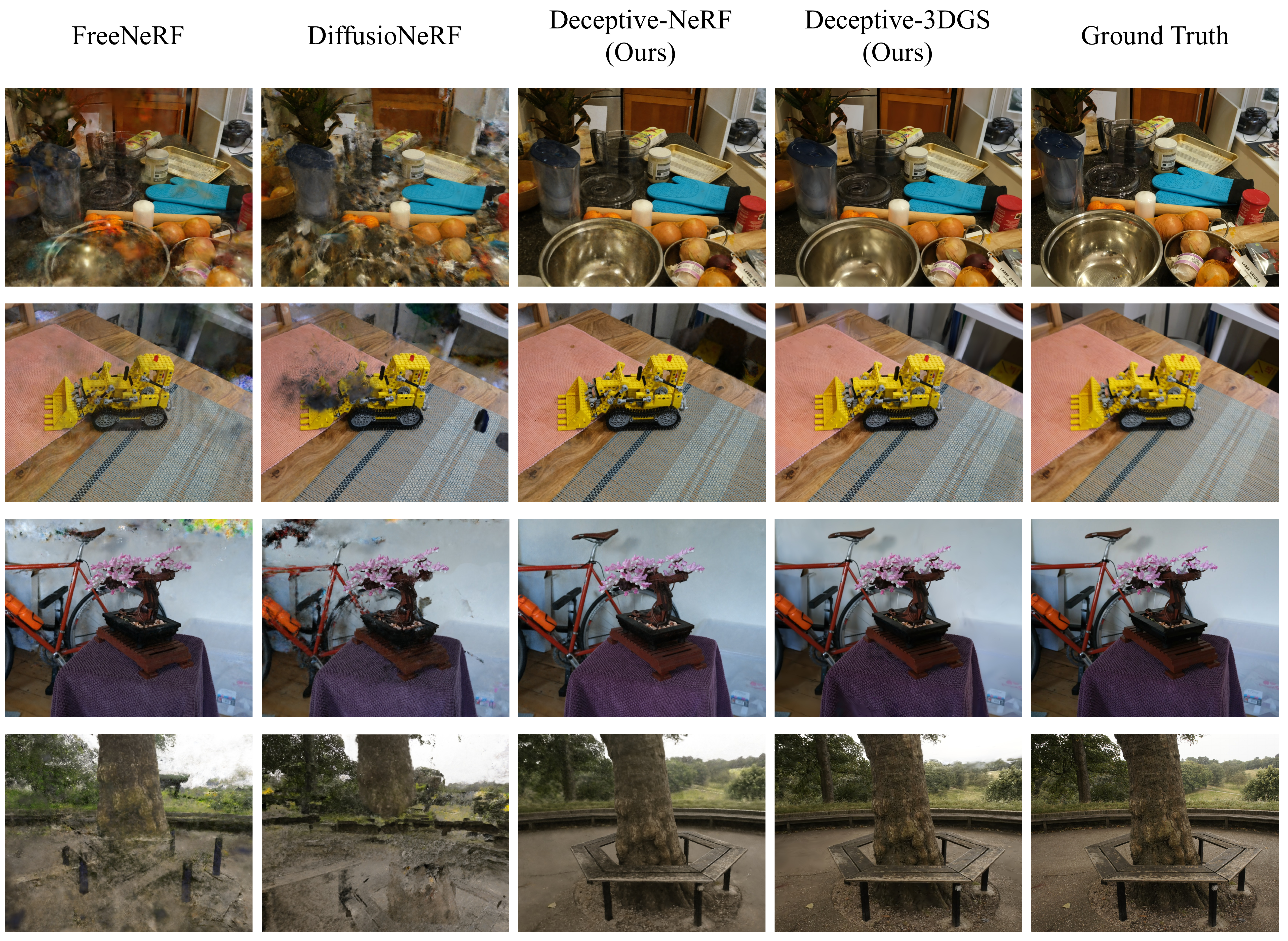}
\end{center}
\vspace{-0.5cm}
\caption{\textbf{Qualitative comparisons of few-view reconstruction on mip-NeRF 360 dataset~\cite{barron2022mipnerf360}.}  For each scene, we reconstruct with 10 input views. Our methods can provide high-quality reconstructions, whereas baseline methods may yield completely unreasonable and incorrect reconstructions.}
\label{fig:360}
\vspace{-0.5cm}
\end{figure}
As illustrated in \Cref{fig:360} and \Cref{fig:hypersim}, our method's synthesized novel views do not produce floating artifacts and achieve better restoration of distant scenes. Even in the challenging 360-degree scenes, our method consistently delivers high-quality reconstructions, outperforming baseline methods that may result in wholly unreasonable and incorrect reconstructions. This showcases our method's advanced few-view reconstruction capabilities over the baselines.

\noindent\textbf{Novel view super-resolution.} We further validate the capability of our method to perform novel-view super-resolution in 360-degree scenes. Utilizing 20 input views and downsampling the input images by a factor of 4 for each scene on the mip-NeRF 360 dataset~\cite{barron2022mipnerf360}, our method, Deceptive-3DGS, achieves high-quality super-resolution at novel viewpoints. It recovers details of objects, such as Lego toys and flower petals, more effectively than competing approaches like FreeNeRF~\cite{yang2023freenerf} and NeRF-SR~\cite{wang2022nerf}.

\begin{figure}[t]
\vspace{-0.5cm}
\begin{center}
    \includegraphics[width=\linewidth]{ 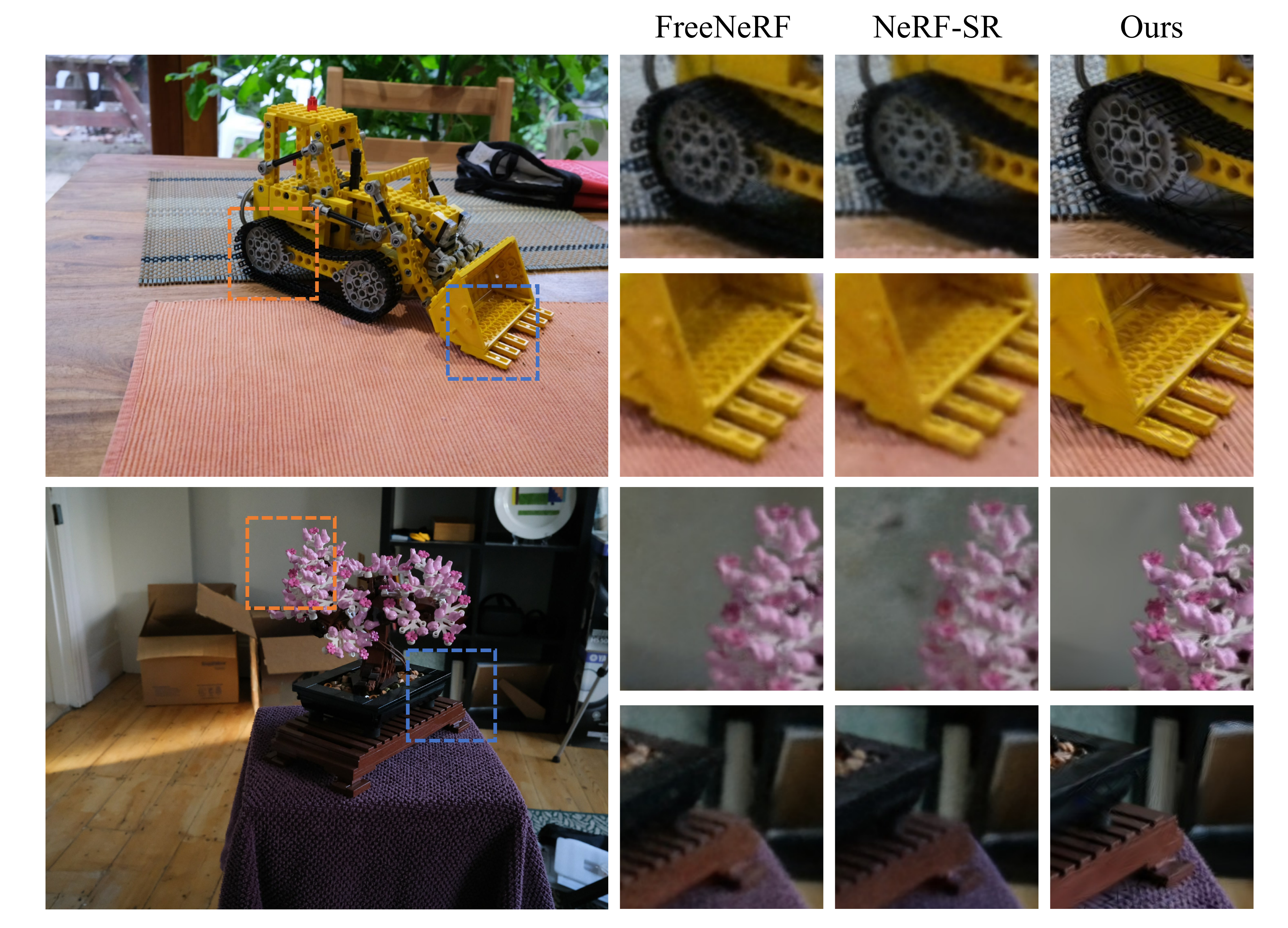}
\end{center}
\vspace{-0.5cm}
\caption{\textbf{Qualitative comparisons of novel view super-resolution on mip-NeRF 360 dataset~\cite{barron2022mipnerf360}.} For each scene, we utilize 20 input views and downsample the input images by a factor of 4. Our method (Deceptive-3DGS) manages to achieve high-quality super-resolution at novel viewpoints, recovering details of objects such as Lego toys and flower petals more effectively than baseline approaches.}
\label{fig:supp_sr}
\end{figure}

\section{Additional Evaluations}
\label{sec:add_eval}
\noindent\textbf{``Regularizer'' v.s. ``view densifier''}
Different from the straightforward utilization of the 2D diffusion model as a ``scorer'' for synthesized novel views to regularize NeRF/3DGS training, our approach uses it to generate pseudo-observations to densify observations.
To better validate the advantages of this choice, we evaluate the reconstruction quality and efficiency between the two methods of using diffusion models to enhance 3D reconstruction.
We experiment on the Hypersim dataset~\cite{roberts2021hypersim} with 10 input views. In \Cref{tab:efficiency}, the ``view densifier'' refers to our proposed Deceptive-NeRF, while the ``regularizer'' denotes a variant of our method that uses a diffusion prior to regularize NeRF training.
Our proposed ``view densifier'' approach outperforms the ``regularizer'' in all metrics of rendering quality. Moreover, our approach achieves nearly ten times faster training speed and increased rendering speed. This improvement is primarily because our method does not require inferring the diffusion model at every training step.

\begin{table}[h]
\centering
\caption{Quantitative comparison of two methods to utilize 2D diffusion models for 3D reconstruction.}
\vspace{-0.25cm}
\label{tab:efficiency}
\resizebox{\linewidth}{!}{
\begin{NiceTabular}{@{}l ccc c c@{}}
\toprule
\thead[l]{Method} & {\thead[c]{PSNR ($\uparrow$)}} & {\thead[c]{SSIM ($\uparrow$)}} & {\thead[c]{LPIPS ($\downarrow$)}} & {\thead[c]{Training Time ($\downarrow$)}} & {\thead[c]{Rendering FPS ($\uparrow$)}} \\
\midrule
``Regularizer'' & 19.31 & 0.710 & 0.253 & 2h40min & 1.1 \\
``View Densifier'' (ours) & \textbf{20.44} & \textbf{0.748} & \textbf{0.173} & \textbf{17min} & \textbf{5.3} \\
\bottomrule
\end{NiceTabular}
}
\vspace{-0.5cm}
\end{table}

\begin{table}[h]
\centering 
    \caption{Quantitative comparison on geometry recovery.}
    \label{tab:geo}
        \begin{tabular}{@{}lcccc@{}}
            \toprule
            & NeRF & FreeNeRF & DiffusioNeRF & Deceptive-NeRF (Ours) \\
            \midrule
            Chamfer Distance ($\downarrow$) & 3.75 & 1.93 & 3.63 & \textbf{1.86} \\
            \bottomrule
        \end{tabular}

\end{table}

\begin{figure}[h]
\centering

\includegraphics[width=0.8\linewidth]{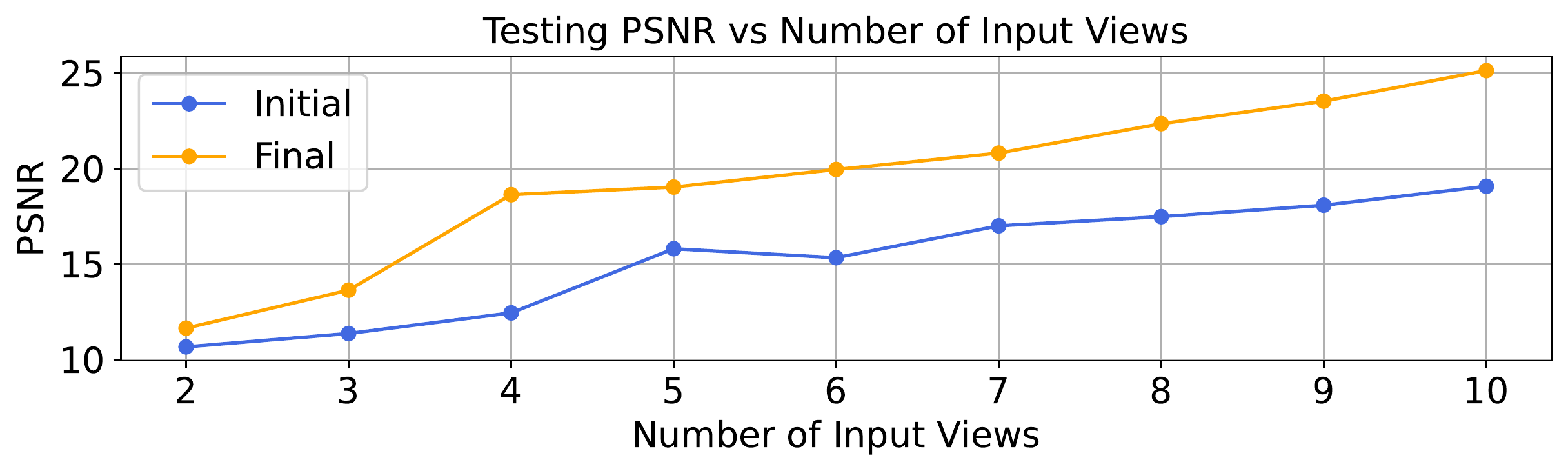}

\caption{Initial and final testing PSNRs across different numbers of input views. }

\label{fig:r1_curve}
\end{figure}

\noindent\textbf{Number of input views.}
We run Deceptive-NeRF on Scene 027\_003 from Hypersim across a range of 2 to 10 input views and examine the testing PSNRs of initial NeRFs and final NeRFs (10x densification). As depicted in \Cref{fig:r1_curve}, our pseudo-observations consistently enhance the reconstruction quality, demonstrating that our method can work effectively with as few as 3 input images.

\noindent\textbf{Geometry recovery.} 
We quantitatively evaluate geometry recovery on scan 65 of the DTU dataset with 9 input views, applying TSDF (truncated signed distance function) Fusion to extract meshes from trained Deceptive-NeRF and baseline methods. In \Cref{tab:geo}, we report their Chamfer distances to the ground truth, and our approach achieves higher reconstruction accuracy than baselines.

\begin{figure}[t]
\begin{center}
    \includegraphics[width=0.7\linewidth]{ 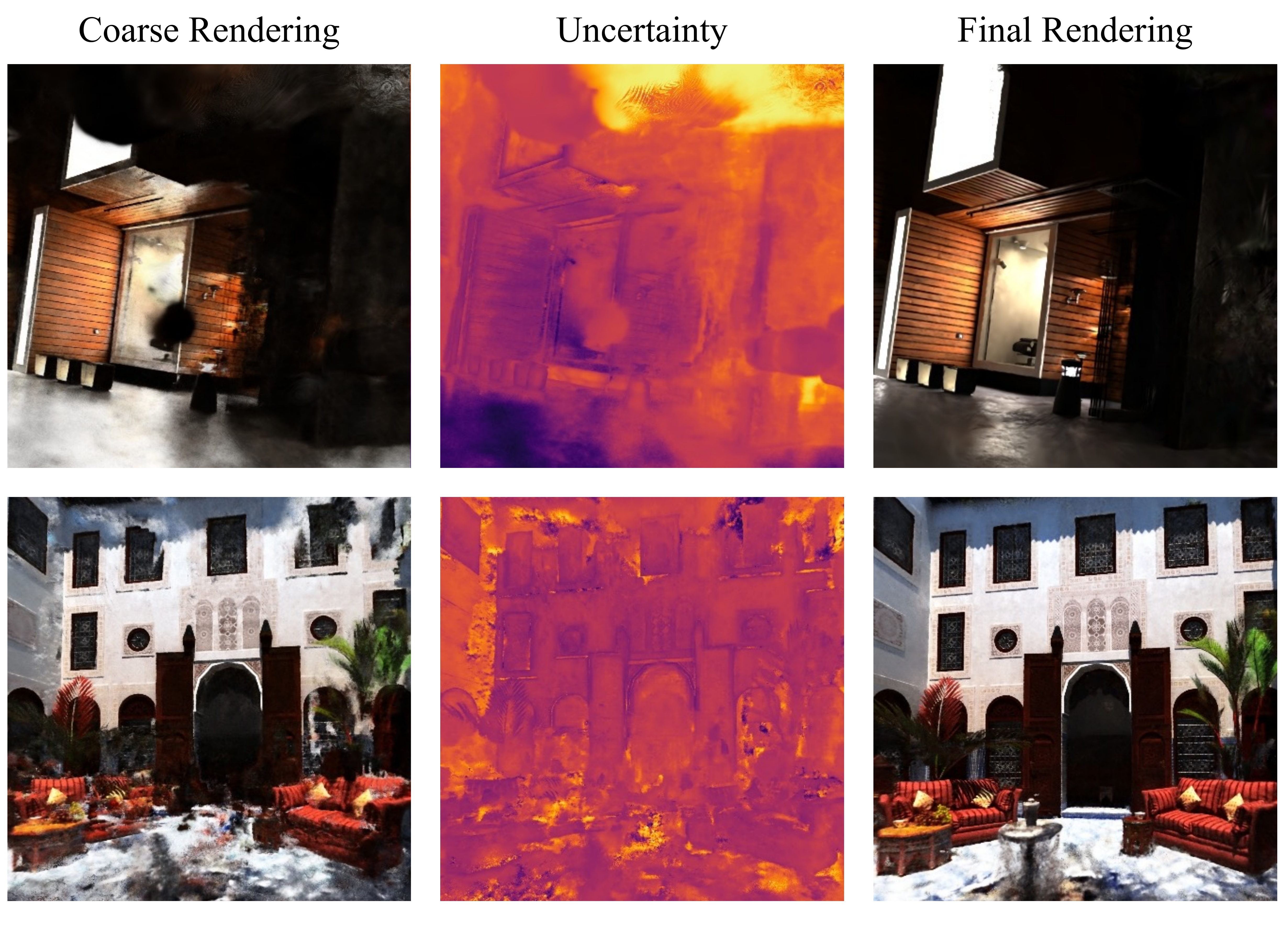}
\end{center}
\vspace{-0.5cm}
\caption{\textbf{Visualizations of uncertainty guidance.} Under the uncertainty measure's guidance, coarse rendering artifacts are eliminated in the pseudo-observations. In uncertainty maps, high uncertainty is indicated by brighter colors, while low uncertainty is shown in darker colors.}
\label{fig:supp_unc}
\end{figure}

\noindent\textbf{Uncertainty guidance.} 
We conducted an ablation study on our uncertainty guidance on the Hypersim dataset~\cite{roberts2021hypersim}.
As shown in \Cref{tab:uncertainty}, without the guidance of our proposed uncertainty measure, there is a significant decline in reconstruction quality. Without uncertainty guidance, the deceptive diffusion model cannot effectively remove artifacts in coarse renderings, leading to inconsistencies between pseudo-observations and input images.
In \Cref{fig:supp_unc}, we visualize the uncertainty map and compare the coarse and final renderings under its guidance. Areas of high uncertainty (highlighted in brighter colors) correspond to artifacts, which are removed in the final rendering.
\begin{table}[h]
\centering
\caption{Quantitative ablation study on the uncertainty guidance.}
\vspace{-0.25cm}
\label{tab:uncertainty}
\resizebox{0.65\linewidth}{!}{
\begin{NiceTabular}{@{}lccc@{}}
\toprule
\thead[l]{Method} & {\thead[c]{PSNR ($\uparrow$)}} & {\thead[c]{SSIM ($\uparrow$)}} & {\thead[c]{LPIPS ($\downarrow$)}} \\
\midrule
w/o uncertainty & 18.61 & 0.703 & 0.28 \\
w/ uncertainty (ours) & \textbf{20.44} & \textbf{0.748} & \textbf{0.173} \\
\bottomrule
\end{NiceTabular}
}
\end{table}

\section{Experimental Details}
\label{sec:details}
We adopt Nerfacto and Splatfacto from NerfStudio~\cite{tancik2023nerfstudio} as the backbones for Deceptive-NeRF and Deceptive-3DGS, respectively, utilizing the default proposal sampling, scene contraction, and appearance embeddings. We randomly initialize the Gaussians for Deceptive-3DGS, without utilizing Colmap point clouds.
We alternately generate pseudo-observations and train scene representations five times, ultimately densifying observations to ten times their original amount.
We randomly sample novel views $\{\phi_{\text{pseudo}}^i\}$ within the bounding box defined by the outermost input cameras.

\end{document}